\documentclass[10pt,twocolumn,letterpaper]{article}

\usepackage[pagenumbers]{wacv} %

\newcommand{\TODO}[1]{\textbf{\color{red}[TODO: #1]}}
\usepackage{booktabs}
\usepackage{multirow}
\usepackage{graphicx}
\usepackage{adjustbox}
\usepackage{colortbl}
\usepackage{amsmath}
\usepackage{algorithm}
\usepackage{algpseudocode}

\AtEndPreamble{
  \crefname{appendix}{Appendix}{Appendices}
  \Crefname{appendix}{Appendix}{Appendices}
}
\definecolor{posgreen}{HTML}{0A7D36}
\definecolor{negred}{HTML}{C22F2F}

\newcommand{\siglip}{SigLIP2}
\renewcommand{\TODO}[1]{}

\newcommand{\platform}{Wix}
\newcommand{\platformfull}{Wix.com}   %
\newcommand{\Platform}{Wix}
\newcommand{\clspfx}{wixui}

\newcommand{\codeurl}{\href{https://wix-incubator.github.io/UI-encoder/}{https://wix-incubator.github.io/UI-encoder/}}

\newif\ifarxiv
\DeclareRobustCommand{\suppref}[1]{\ifarxiv\Cref{#1}\else The supplementary material\fi}
\DeclareRobustCommand{\inmain}{\ifarxiv\else{} in the main paper\fi}

\arxivtrue  %
\usepackage[accsupp]{axessibility}  %

\definecolor{wacvblue}{rgb}{0.21,0.49,0.74}
\usepackage[pagebackref,breaklinks,colorlinks,allcolors=wacvblue]{hyperref}

\def\wacvPaperID{26} %
\def\confName{WACV}
\def\confYear{2027}

\title{Masking by Design: Component-Conditioned Pretraining for User Interfaces}

\author{Shon Otmazgin$^{*}$ \quad
Shahar Nahum$^{*}$ \quad
Adi Dick-Charnilas\\
Gershon Goland \quad Amit Israeli \quad Gilad Barkan\\
Wix.com, Israel\\
{\tt\small \{shono, shaharn, adidi, gershong, amitisr, giladba\}@wix.com}
}

\begin{document}
\maketitle
{\renewcommand{\thefootnote}{*}\footnotetext{Equal contribution.}}

\begin{abstract}
Web user interfaces are composed rather than captured: a page is built from components, and recovering those components from pixels underlies design tooling, accessibility, UI testing and software agents.
Once rendered, the page holds two representations: the screenshot an encoder sees, and the structured document it was drawn from, which names those components and fixes the box each one occupies. Prior work has read that document as context, or predicted it back as a target. We propose \emph{component-conditioned masking}, which puts it to a different use: it authors the mask for pretraining the image encoder. The result is masking by design: part of every component in the twelve classes we target is hidden, regardless of its size, position, or frequency, so each one becomes a prediction target.
Our corpus is $136{,}277$ rendered pages from \platformfull{}, a website-building platform where a single vocabulary names every component. We continue pretraining a \siglip{} encoder with I-JEPA under this policy, and measure it with a linear probe that predicts each component's class: macro F1 rises from $61.5$ to $66.0$. To isolate the mask we pretrain again with random masking, changing nothing else: the difference is $1.3$ macro F1 across three seeds, and it is gained on the components a random draw under-covers---five of the twelve classes---where ours masks more than half of the area of every class. Our mask improves all twelve classes we evaluate, while the random draw leaves some below the frozen \siglip{} encoder. We release the code, the pretrained encoders, and $9{,}000$ pages---the sealed test split and the split our probe trains on---with every component's class and box.\footnote{Code, data and encoders: \codeurl.}
\end{abstract}

\section{Introduction}
\label{sec:intro}

\begin{figure}[t]
  \centering
  \includegraphics[width=\linewidth]{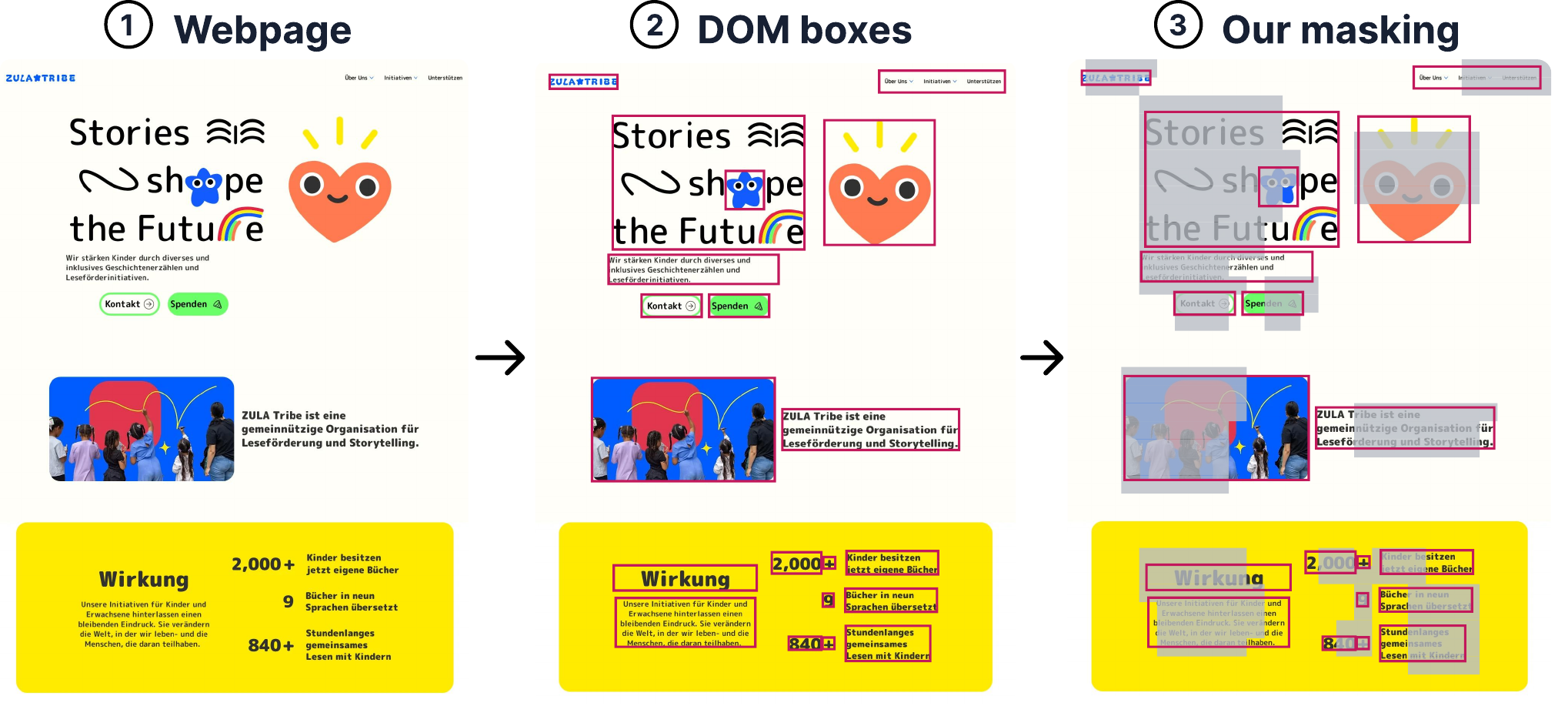}
  \caption{\textbf{Component-conditioned masking}, on a real page from our corpus.
  \emph{(1) Webpage}: the rendered screenshot, the only input the encoder receives.
  \emph{(2) DOM boxes}: the box each component occupies, taken from the page's own structured document and outlined here in magenta.
  \emph{(3) Our masking}: the mask is placed inside those boxes---part of every component.
  Gray marks the masked target patches; what stays visible is the context they are predicted from.}
  \label{fig:pipeline}
\end{figure}

\begin{figure*}[t]
  \centering
  \includegraphics[width=0.70\textwidth]{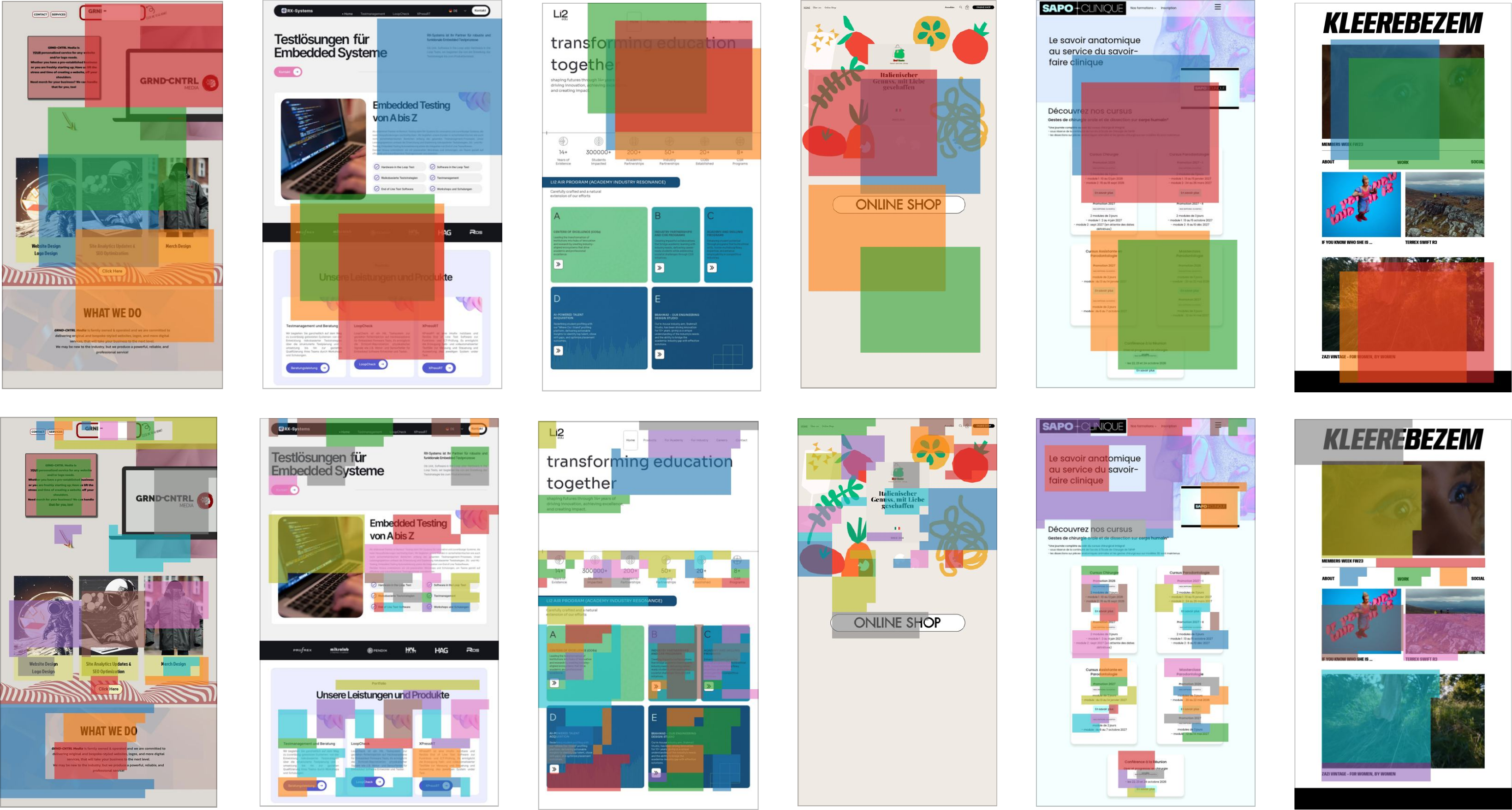}
  \caption{\textbf{The same six pages under both policies.}
  \emph{Top}: random masking is blind to the page, so components go untouched on one page and vanish whole on the next.
  \emph{Bottom}: ours places the mask inside the boxes the page's document records, part of every component, on every page.
  Colors mark the target blocks to be predicted.}
  \label{fig:examples}
\end{figure*}

User interfaces are the language through which people interact with software. They are constructed from buttons, menus, text, images, input fields, and other components, arranged to convey information and afford interaction. Machines are increasingly asked to read this language: to anticipate which component a user will act on next~\cite{he2021actionbert}, to describe what a screen contains and does~\cite{wang2021screen2words}, to retrieve relevant interfaces or UI components using visual or textual queries~\cite{bunian2021vins, deka2017rico}, and to ground a natural-language instruction in the relevant on-screen content~\cite{bai2021uibert}. All of these start from the same input: the rendered screen. The screen arrives as raw pixels, and the pixels must become a representation that preserves how the interface looks, what it is built from, and how it is arranged. This is the image encoder's job---and what its representation misses, no later stage recovers.

Masked prediction, a self-supervised recipe~\cite{he2022mae,assran2023ijepa}, trains such an image encoder by masking a \emph{target} region of the image and predicting it from the remaining \emph{context}. It yields representations that transfer broadly across vision tasks~\cite{bao2022beit,wei2022maskfeat,zhou2022ibot}. But the recipe says nothing about where the mask falls, so by default it falls at random. Where it falls is not a neutral detail: it changes what the encoder is encouraged to learn. Later work therefore aims the mask at the image's parts, estimating them from the pixels---with a segmentation model~\cite{li2022semmae}, a teacher's attention~\cite{kakogeorgiou2022attmask,he2026dseqjepa}, or a mask learned during training~\cite{shi2022adios}---since a natural image carries no record of what its parts are or where they lie.

In web pages, this information is already embedded in the rendering process. Before a webpage becomes pixels, it is a structured document---the Document Object Model (DOM)~\cite{whatwg2026dom}---among whose nodes are the page's components, each assigned a box by layout. A webpage therefore has two simultaneous forms: the screenshot image presented to the encoder and the structured skeleton from which it was rendered. Prior work has used structure as context supplied to the model~\cite{li2021screen2vec,he2021actionbert,bai2021uibert}, or as a prediction target reconstructed from pixels~\cite{lee2023pix2struct}. We propose \emph{component-conditioned masking} (\cref{fig:pipeline}), a policy that puts it to a different use: the DOM decides where the pretraining mask falls in the screenshot, and the model predicts the hidden visual content. The mask is drawn inside the page's components, part of each; since only masked patches are predicted, every component becomes a prediction target---whatever its size, position, or frequency. A random draw makes no such promise: which components it reaches is a matter of where its blocks happen to land (\cref{fig:examples}).

To train and evaluate the policy, we crawled and rendered $136{,}277$ webpages from \platformfull{}, a website-building platform that names every component from one vocabulary, so each screenshot comes with precise boxes and classes. We continue pretraining a public \siglip{} encoder~\cite{tschannen2025siglip2} with I-JEPA~\cite{assran2023ijepa} on the training split, and measure it with a linear probe that predicts a component's class, one of twelve, from the encoder representations. No class is predicted before this point; the mask itself is placed from the DOM boxes. We release the sealed test split together with the split the probe trains on---$9{,}000$ pages, every component with its class and box---so the same evaluation can be run on any encoder.

Our experiments show that our masking policy raises macro F1 over the frozen \siglip{} encoder, from $61.5$ to $66.0$ (\cref{tab:headline}). Most of that is the domain: pretraining on webpages moves any encoder toward them. To isolate the mask we pretrain again with random masking as the alternative, changing nothing but the mask. Averaged over the twelve classes, ours is ahead of random masking by $1.3$ points ($p{=}0.045$, three seeds). Examining the classes individually reveals a sharper pattern. Ours is the only policy that improves every one of the twelve classes (\cref{tab:headline})---random masking leaves some worse than the frozen encoder. We analyze where these gains come from and find a clear signal: coverage---the share of a class's patches the mask hides, and therefore turns into prediction targets---tracks the largest improvements. Random masking distributes it unevenly, hiding more than half of \emph{text}'s patches but under a tenth of \emph{hamburger}'s. Ours instead hides more than half of every class's patches, so every component contributes to the learning signal (\cref{tab:mechanism}). Component coverage added and F1 gained are correlated across the twelve classes ($r{=}0.72$). We further ask whether balancing the loss across classes can substitute for coverage. Partly. A balanced loss lifts random masking above our mask in macro F1 (\cref{tab:oursplus}), yet ours still leads on every thinly covered class, those random masking covers least. Reweighting can only amplify the patches a mask selects; it cannot add the ones it misses. The two are complementary, and together---our masking with the balanced loss---they give the largest gain.

\paragraph{Contributions.}
\begin{itemize}
  \item \textbf{Component-conditioned masking.} The DOM places the pretraining mask by design: part of every component in the twelve classes we target. The image encoder learns by predicting each component's hidden content back.
  \item \textbf{Gains track coverage, not mask budget.} Per class, the F1 gained follows the coverage added (correlation $r{=}0.72$), not the amount masked: ours hides $34\%$ of the page against the alternative's $50\%$.
  \item \textbf{Code, weights and a sealed benchmark.} We release the pretrained encoders, the code, and $9{,}000$ pages---the sealed test split and the split the probe trains on---with every component's box and its class from a single vocabulary. Any encoder can then be scored on component-level UI representation exactly as ours is.
\end{itemize}

\section{Background}
\label{sec:background}

We pretrain image encoders on screenshots of web pages. A web page has two coupled representations: a structured document and a visual image. First, as a structured document, the browser maintains the Document Object Model (DOM)~\cite{whatwg2026dom}: a tree of the nested nodes a web page is built from. Each node records what it is---its tag, its classes, its content---and, once the browser lays the page out, the \emph{bounding box} it occupies on the screen. The tree also captures hierarchy---the links of a menu bar are children of the menu's node. Among these nodes are the page's \emph{components}---menus, buttons, headers, paragraphs of text; \cref{sec:dataset} describes which nodes we treat as components and how we extract them. Second, as a visual image, the browser renders the page into pixels, and a screenshot of them is our input image. Crucially, the two representations are spatially aligned: the bounding box the DOM records for a component is the region the browser lays that component out in, and so locates it in the screenshot.

\paragraph{Masked prediction pretraining.}
Masked prediction hides part of an image and asks a model to predict the hidden part from the rest~\cite{pathak2016context,bao2022beit,xie2022simmim,hondru2025mimsurvey}. We adopt its latent-space form, I-JEPA~\cite{assran2023ijepa}, which makes that prediction in feature space.\footnote{MAE~\cite{he2022mae} is the pixel-space counterpart.} Given an image $x$, a vision transformer encoder, ViT~\cite{dosovitskiy2021vit}, splits it into $N$ patches $p_1,\dots,p_N$. A \emph{masking policy} $\pi$ then splits the positions $\{1,\dots,N\}$ into two disjoint sets: a \emph{target} set $T$ of \emph{masked} positions, to be predicted, and a \emph{context} set $C$ of \emph{visible} ones, seen and used as context to predict $T$.\footnote{Some settings also allow a third set $D$: patches dropped from $C$, neither predicted nor seen. We do not use one.} Two encoders then read the patches. The \emph{context encoder} $f_\theta$ sees only those at the positions in $C$ and embeds them, $z=f_\theta(\{p_k\}_{k\in C})$. The \emph{target encoder} $f_{\bar\theta}$ sees all $N$ patches, and its features supply the targets: $t_k=f_{\bar\theta}(\{p_n\}_{n=1}^{N})_k$ for $k\in T$. The target encoder is either held fixed or maintained as an exponential moving average of the context encoder. A narrow ViT \emph{predictor} $g_\psi$ bridges the two encoders: given $z$ and the positions in $T$, it estimates the targets, $\hat t_k=g_\psi(z,T)_k$ for $k\in T$. Training minimizes the feature-space distance over the target set alone,
\begin{equation}
  \mathcal{L}=\frac{1}{|T|}\sum_{k\in T} d\big(\hat t_k,\,t_k\big),
  \label{eq:loss}
\end{equation}
with $d(\cdot)$ the smooth-$\ell_1$ distance. So the only learning signal is how well the target patches are recovered from the context ones.

\paragraph{Where to mask.}
By \cref{eq:loss}, everything the model learns comes through the mask. How much it hides is the \emph{mask ratio}, $|T|/N$; where it falls is our subject. Common practice draws the mask at random---scattered patches~\cite{he2022mae} or a handful of rectangular blocks~\cite{assran2023ijepa}---blind to what the image contains. Other policies \emph{estimate} the mask from the image~\cite{li2022semmae,kakogeorgiou2022attmask} or \emph{learn} it during training~\cite{shi2022adios,wang2023hpm}; the full survey is in \cref{sec:related}. \emph{Ours} is \emph{authored}: the DOM already records the decomposition, so the mask is built from structure the web page supplies.

\section{Component-Conditioned Masking}
\label{sec:masking}

A web page carries components of many kinds, sizes and positions: a header across the top, a menu inside it, a login bar in the corner, images and text below. Our policy takes the masked-prediction recipe of \cref{sec:background} from the page level down to the component level: hide part of every component, and predict it from what remains---the small and peripheral no less than the large and central. Masking a component puts both of its parts to work: what stays in context is encoded, and what is hidden must be predicted from that encoding. The two are complementary---the visible part is the representation, the hidden part is what that representation has to account for---so we design the mask to leave every component both.

\paragraph{The masking policy.}
\label{sec:masking:set}
We traverse the page's components $E$, drawing a random block inside each; their union is the page's mask. Let $P_e$ be the patch positions component $e\in E$ covers, which we split into two disjoint sets, $P_e=T_e\cup C_e$. The hidden \emph{target part} $T_e$ is a contiguous block along the component's longer axis, placed at random and sized to a share $\rho_e\sim\mathcal{U}(a,b)$ of its patches. The \emph{context part} $C_e$ is what remains, and stays visible to the encoder. Every component draws its own $\rho_e$ from the same band $(a,b)$, whatever its size and wherever it is located in the page, so every component takes part in the mask. The page's mask is then defined as
\begin{equation}
  T \;=\; \bigcup_{e\in E} T_e
  \qquad ; \qquad C \;=\; \{1,\dots,N\}\setminus T
  \label{eq:mask}
\end{equation}
So $T$ lies entirely on components, and $C$ holds their context parts together with the page's non-component area.

\paragraph{Nesting.}
\label{sec:masking:nesting}
Some components sit inside others---a menu bar contains its links---so one component's target block can fall inside another. Children are masked first, and what a child hides counts toward its parent's share; the parent's target block then goes around what its children keep as context, never over it. So every component, parent or child, keeps a context part of its own.

\cref{fig:pipeline} demonstrates the masking policy's result on a real page. Each component is partly hidden and nothing else is: the non-component areas are never masked, never predicted, only ever context. The DOM's only role is to decide where the mask falls. Everything else about training is unchanged (\cref{sec:background}). 

\section{Dataset}
\label{sec:dataset}

Our corpus holds 136{,}277 pages from live websites, all built with \platformfull{}, a commercial website-building platform. Each page goes through the same pipeline: render it, cut it at a section boundary, then capture the screenshot and record the components inside the cut.

\paragraph{Crawling.}
\label{sec:masking:extract}
The crawler loads each URL in a headless browser at a width of 1280 pixels and waits for the page to finish rendering. Using the page's DOM it then records every component on the page together with the bounding box it occupies, and takes a screenshot. Pages are sampled at random, one per site---its homepage.

\paragraph{What counts as a component.}
A page's DOM is a tree of nested nodes. A few declare what they are---a
\texttt{\textless button\textgreater} is a button---but most are generic
containers and wrappers, so the tree by itself does not say which nodes are
the page's components. Two sites can name and build the same
component differently, so labels recovered from the open web stay coarse;
WebUI~\cite{wu2023webui}, which pairs webpage screenshots with their DOM at
scale, draws on that web. Our pages come from a single platform,
\platform{}, which stamps every component it
renders with its own class token, such as \texttt{\clspfx-button},
\texttt{\clspfx-dropdown-menu} or \texttt{\clspfx-rich-text}: one vocabulary,
applied by the platform itself, the same on every page. A DOM node is a
component if it carries a \texttt{\clspfx-*} token---that is the whole
definition. \Cref{tab:inventory} lists the component tokens this paper uses.

\paragraph{Capturing a scrolling page.}
There is more than one way to screenshot a web page: the whole page, the viewport, or a fixed crop. Capturing the whole page can result in an image five, ten, even twenty screens tall, and the encoder's patch budget must then stretch across all of it. The viewport and a fixed crop both end at an arbitrary line, cutting through whatever component sits there. We cut the page at a section boundary instead. \Platform{} marks sections as it marks components, so we know where each one ends: we set a target height of 2160 pixels and cut the screenshot at the section boundary closest to it. Screenshots therefore vary in height,\footnote{\suppref{sec:appendix:shots} gives the distribution of screenshot dimensions.} and a section that runs long carries the cut past the target. A page in our corpus consists of the screenshot and the components inside it; anything below the cut is dropped, so every component we record appears in the screenshot.

\begin{table}[t]
  \centering
  \caption{Component inventory: the \texttt{\clspfx-*} tokens we keep and the classes they map to. \emph{pages}: share of the $136{,}277$ pages carrying the token. \emph{inst.\ (K)}: instances, in thousands. \emph{med.\ \%}: median instance size as a share of the page's patch grid (the median page is ${\sim}1{,}000$ patches of 16\,px, so $1\%\approx10$ patches). A page carries $15.6$ whitelisted instances on average. \suppref{sec:appendix:taxonomy} lists the full vocabulary, kept and ignored alike.}
  \label{tab:inventory}
  \scriptsize
  \setlength{\tabcolsep}{3pt}
  \begin{tabular}{@{}llrrr@{}}
    \toprule
    component token & class & pages & inst.\ (K) & med.\ \% \\
    \midrule
    \texttt{rich-text} & \multirow{2}{*}{text} & 93\% & 990 & \multirow{2}{*}{2.0} \\
    \texttt{collapsible-text} &  & 1\% & 3 &  \\
    \addlinespace[2pt]
    \texttt{image} & \multirow{3}{*}{image} & 84\% & 406 & \multirow{3}{*}{4.0} \\
    \texttt{video-box} &  & 11\% & 19 &  \\
    \texttt{video-player} &  & 3\% & 5 &  \\
    \addlinespace[2pt]
    \texttt{horizontal-menu} & \multirow{2}{*}{menu} & 54\% & 76 & \multirow{2}{*}{2.8} \\
    \texttt{vertical-menu} &  & 1\% & 1 &  \\
    \addlinespace[2pt]
    \texttt{text-input} & \multirow{2}{*}{input-field} & 5\% & 12 & \multirow{2}{*}{1.4} \\
    \texttt{search-bar} &  & 3\% & 4 &  \\
    \addlinespace[2pt]
    \midrule
    \texttt{button} & button & 69\% & 256 & 0.8 \\
    \texttt{vector-image} & vector-image & 42\% & 149 & 0.4 \\
    \texttt{header} & header & 85\% & 117 & 7.5 \\
    \texttt{repeater} & repeater & 22\% & 38 & 18 \\
    \texttt{dropdown-menu} & dropdown-menu & 19\% & 26 & 3.2 \\
    \texttt{login-social-bar} & login-social-bar & 12\% & 16 & 0.6 \\
    \texttt{hamburger-open-button} & hamburger & 6\% & 9 & 0.4 \\
    \texttt{text-marquee} & text-marquee & 4\% & 7 & 4.4 \\
    \addlinespace[1pt]
    \emph{class taxonomy} & & & \emph{2{,}131}\,(51\%) & \emph{2.0} \\
    \midrule
    168 others & non-component & --- & 2{,}008\,(49\%) & --- \\
    \midrule
    \emph{total} & & & \emph{4{,}139} & \\
    \bottomrule
  \end{tabular}
\end{table}

\paragraph{Class taxonomy.}
In \cref{tab:inventory} we fix the components this paper uses: seventeen tokens, mapped into twelve classes. From here on, a \emph{component} is an instance of one of them, and the mask (\cref{sec:masking}) falls only on these. The other $168$ tokens are treated as non-component area of the screenshot. Several components share a class: \texttt{\clspfx-rich-text} and \texttt{\clspfx-collapsible-text} are both mapped to \emph{text}. Most of what we ignore are containers such as \texttt{\clspfx-box} and \texttt{\clspfx-section}, which arrange other components rather than carrying content of their own; the rest are rare. Not every container is ignored, though: \emph{repeater} and \emph{menu}, for instance, hold other components too, and we keep them.

\paragraph{Pages and splits.}
The 136{,}277 pages split into four sets, each page in exactly one: train (125{,}277), seed (4{,}000), validation (2{,}000), and held-out test (5{,}000). The test split is sealed and read once. We release it together with the seed split, which trains the classifier used for evaluation.\footnote{\suppref{sec:release} carries the ethics statement, including the takedown mechanism.}

\section{Experiments}
\label{sec:experiments}

\paragraph{Setup.}
\siglip{}-so400m~\cite{tschannen2025siglip2} is our backbone: every run continues the same public weights in their native-resolution form, NaFlex, capped at 1{,}024 patches per page. The target encoder, $f_{\bar\theta}$, is a frozen copy of those weights; the context encoder, $f_{\theta}$, starts from them; and the predictor, $g_\psi$, is a lightweight six-layer ViT (six heads, width 384), randomly initialized. Freezing the target encoder follows masked feature-distillation practice, where the targets come from a strong pretrained encoder held fixed~\cite{fang2023eva,hou2022milan,liu2022dbot}. All runs share one recipe: AdamW ($\beta_1{=}0.9$, $\beta_2{=}0.999$), a cosine learning rate peaking at $5{\times}10^{-5}$ after a 1{,}000-step warm-up, weight decay 0.05, gradient clipping at 1.0, and 17{,}000 steps at an effective batch of 64 pages---about nine epochs. The context encoder's learning rate is held at zero for the first 500 steps, so the predictor fits before the encoder starts to move. We filter the train set to pages whose components cover at least $10\%$ of the page---below that there is too little for a component mask to hold---which takes it from $125{,}277$ to $118{,}583$ pages.

\paragraph{What we compare.}
Four encoders meet in the headline comparison of \cref{tab:headline}. The frozen \emph{\siglip{}}-so400m encoder, with no further training, is the baseline the rest are measured against. We then consider two random masking policies. The first, \emph{random}, draws its targets as I-JEPA does:\footnote{I-JEPA also crops the context to $85$--$100\%$ of the image; we keep the whole context.} four target blocks per page, each covering $15$--$20\%$ of it, at aspect ratio $0.75$--$1.5$. The second, \emph{random$^{+}$}, replaces them with sixteen target blocks of $2$--$8\%$ each---still a random draw, but matched to the corpus inventory (\cref{tab:inventory}): a page carries $15.6$ components on average, and $2$--$8\%$ spans the typical range of their median sizes. Last is \emph{ours}, which implements the masking policy of \cref{sec:masking} at the band $(0.5,0.7)$. The three continued-pretrained models share the same starting weights, recipe, data and seeds; any two differ by their mask alone.

\paragraph{How we compare.}
We measure how well each encoder represents components with the standard linear-probing protocol: a linear head predicts each component's class from the frozen features, so whatever it reads must already be in the representation. Class labels enter the pipeline here for the first time as targets. No class is predicted during pretraining: the mask is placed from the DOM boxes of the whitelisted components, and where the loss is balanced (\cref{tab:oursplus}) class labels only weight patches. The encoder reads the full page, unmasked; an instance's feature is the mean of the patch features its box overlaps. A linear probe---a twelve-way classifier---is trained on those features, on instances drawn from the seed split, and tuned by cross-validation within it; the sealed test is read once at the end. We report per-class F1 first, because the question is which components improve, and macro F1 as the aggregate.

\begin{table}[!t]
  \small
  \centering
  \caption{Per-class F1 on the test split, linear probe, mean over three pretraining seeds. The \siglip{} column is the frozen encoder's absolute F1; each $\Delta$ is the change from it under continued pretraining with that mask. Bold marks the best mask per class. \emph{Ours} masks at the band $(0.5,0.7)$, selected on the validation split and not revisited. Seed standard deviation of macro F1: ours $\pm0.21$, random$^{+}$ $\pm0.51$, random $\pm0.71$; for the gain over random$^{+}$, a paired $t$-test across the three seeds gives $p{=}0.045$.}
  \label{tab:headline}
  \setlength{\tabcolsep}{5.5pt}
  \begin{tabular}{@{}lrrrr@{}}
    \toprule
    & \siglip{} & \multicolumn{3}{c}{$\Delta$\,F1 over\ \siglip{}} \\
    \cmidrule(lr){3-5}
    class & F1 & random & random$^{+}$ & ours \\
    \midrule
    dropdown-menu    & 53.2 & +10.9 & +11.3 & \textbf{+14.0} \\
    menu  & 65.6 & +4.8 & +6.9 & \textbf{+9.1} \\
    header           & 77.8 & +4.2 & +4.7 & \textbf{+7.5} \\
    vector-image     & 53.4 & +1.0 & +1.6 & \textbf{+3.3} \\
    text             & 82.7 & +2.1 & +2.9 & \textbf{+3.2} \\
    button           & 78.8 & 0.0 & +1.2 & \textbf{+2.1} \\
    login-social-bar & 62.1 & -6.7 & -4.1 & \textbf{+2.0} \\
    hamburger        & 62.5 & -2.6 & -0.9 & \textbf{+1.1} \\
    text-marquee     & 35.4 & +3.9 & \textbf{+6.2} & +5.1 \\
    repeater         & 19.3 & +2.9 & \textbf{+3.4} & +2.5 \\
    image            & 71.1 & +1.7 & \textbf{+2.7} & +2.3 \\
    input-field      & 75.9 & -2.4 & \textbf{+2.0} & +1.7 \\
    \midrule
    macro F1         & 61.5 & +1.7 & +3.2 & \textbf{+4.5} \\
    \bottomrule
  \end{tabular}
\end{table}

\paragraph{Performance.}
Continued pretraining with our policy raises macro F1 from $61.5$ to $66.0$ (\cref{tab:headline}). The two random variants raise the average as well, but not every class with it: under \emph{random}, three classes end below the frozen encoder---input-field by $2.4$ points, hamburger by $2.6$, login-social-bar by $6.7$; \emph{random$^{+}$} recovers input-field and narrows the other two, but leaves both short of it, at $-0.9$ and $-4.1$. Under \emph{ours} every class improves, from $+14.0$ (dropdown-menu) down to $+1.1$ (hamburger), and it is the best of the three in eight of the twelve, with \emph{random$^{+}$} ahead in the remaining four by at most $1.1$. Those per-class margins average to $+1.3$ macro F1 over \emph{random$^{+}$} ($+1.9$, $+1.1$, $+1.0$ on the three seeds) and $+2.8$ over \emph{random}.

\begin{figure}[!t]
  \centering
  \includegraphics[width=\columnwidth]{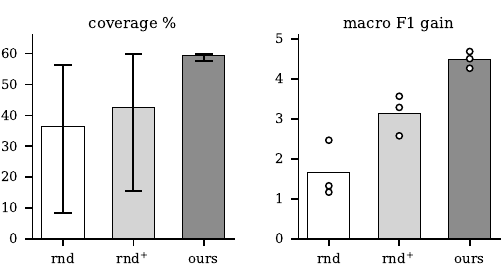}
  \caption{Coverage and the macro F1 gain that follows it. \emph{coverage}: the share of a class's patches the mask hides. Left: bars are the mean over the twelve classes, capped lines their full range. Right: the bar is the mean over three pretraining seeds, the dots those seeds. Coverage on the training pages, macro F1 on the test split.}
  \label{fig:aggregate}
\end{figure}

\begin{table}[!t]
  \footnotesize
  \centering
  \caption{Coverage and gain per class; \emph{random} is the baseline throughout. \emph{coverage}: the share of a class's patches the mask hides---random's as a percentage, the others as multiples of it. $\Delta$F1: change in test F1. \emph{corr.\ coverage/F1}: correlation between coverage added and F1 gained ($n{=}12$; $p$-values $0.87$ and $0.009$). Group means in bold.}
  \label{tab:mechanism}
  \setlength{\tabcolsep}{3.5pt}
  \begin{tabular}{@{}lrrrrr@{}}
    \toprule
    & \multicolumn{3}{c}{coverage} & \multicolumn{2}{c}{$\Delta$F1 vs.\ random} \\
    \cmidrule(lr){2-4} \cmidrule(lr){5-6}
    class & random & random$^{+}$ & ours & random$^{+}$ & ours \\
    \midrule
    \multicolumn{4}{@{}l}{\emph{Thinly covered}} & \textbf{+1.5} & \textbf{+4.6} \\
    \addlinespace[1pt]
    hamburger        & 8\%  & $\times$1.9 & $\times$7.0 & +1.7 & +3.7 \\
    login-social-bar & 11\% & $\times$1.8 & $\times$5.0 & +2.6 & +8.7 \\
    header           & 20\% & $\times$1.5 & $\times$3.1 & +0.5 & +3.3 \\
    menu             & 24\% & $\times$1.4 & $\times$2.5 & +2.1 & +4.3 \\
    dropdown-menu    & 27\% & $\times$1.4 & $\times$2.2 & +0.4 & +3.1 \\
    \midrule
    \multicolumn{4}{@{}l}{\emph{Well covered}} & \textbf{+1.5} & \textbf{+1.6} \\
    \addlinespace[1pt]
    text-marquee     & 44\% & $\times$1.1 & $\times$1.4 & +2.3 & +1.2 \\
    input-field      & 47\% & $\times$1.1 & $\times$1.3 & +4.4 & +4.1 \\
    button           & 48\% & $\times$1.1 & $\times$1.2 & +1.2 & +2.1 \\
    vector-image     & 49\% & $\times$1.1 & $\times$1.2 & +0.6 & +2.3 \\
    repeater         & 50\% & $\times$1.1 & $\times$1.2 & +0.5 & $-$0.4 \\
    image            & 52\% & $\times$1.1 & $\times$1.2 & +1.0 & +0.6 \\
    text             & 56\% & $\times$1.1 & $\times$1.1 & +0.8 & +1.1 \\
    \midrule
    corr.\ coverage/F1 & & & & $r{=}0.05$ & $r{=}0.72$ \\
    \bottomrule
  \end{tabular}
\end{table}

\paragraph{Coverage and gain.}
One statistic explicitly discriminates the two random draws from ours: per-class coverage---the share of each class the mask hides. \cref{fig:aggregate} traces it in aggregate. From \emph{random} to \emph{random$^{+}$}, mean class coverage rises from $36$ to $42\%$ and macro F1 rises with it, $+1.7$ to $+3.2$; under \emph{ours} the whole distribution moves, every class landing at $58$--$60\%$ where the random draws leave them anywhere from $8$ to $60$, and macro F1 reaches $+4.5$ with the seed spread tightening from $\pm0.71$ to $\pm0.21$.

Under \emph{random$^{+}$}, raising coverage already comes with a gain; what it cannot give is a per-component coverage guarantee---exactly what \emph{ours} provides. \cref{tab:mechanism} shows how the gains are distributed, class by class; every number in it is measured against \emph{random}. Sorted by random's coverage, the twelve classes split cleanly in two: seven it already covers at $44$--$56\%$, and five it leaves at $8$--$27\%$. \emph{Random$^{+}$} is indifferent to the split: it gains the same $+1.5$ in both groups, even though it adds more coverage to the \emph{thinly covered} one ($\times1.4$--$1.9$ against $\times1.1$). \emph{Ours} is not: it matches \emph{random$^{+}$} on the \emph{well covered} group at $+1.6$ and departs on the \emph{thinly covered} one, raising coverage $\times2.2$--$7.0$ by construction and gaining $+4.6$. The two policies differ in how they allocate coverage---\emph{random$^{+}$} by chance, ours by design---and the correlations state it: per class, gain and added coverage are uncorrelated under \emph{random$^{+}$} ($r{=}0.05$) and highly correlated under ours ($r{=}0.72$, $p$-value $0.009$). The gain follows coverage where it is aimed.

\begin{table}[!t]
  \small
  \centering
  \caption{Closing the page-level budget gap to the random draws, two ways: widen the band, or top up with non-component blocks. Masked share of the page measured on the training pages, where the mask operates, split into \emph{components} and \emph{non-comp} area. $\Delta$mF1: macro F1 gain over frozen \siglip{} on the validation split.}
  \label{tab:ladder}
  \setlength{\tabcolsep}{6pt}
  \begin{tabular}{@{}lrrrcr@{}}
    \toprule
    & \multicolumn{3}{c}{masked \% of the page} & & \\
    \cmidrule(lr){2-4}
    setting & components & non-comp & total & & $\Delta$mF1 \\
    \midrule
    \multicolumn{6}{@{}l}{\emph{widening the band}} \\
    $(0.4,0.6)$          & 28 & 0  & 28 & & +3.6 \\
    $(0.5,0.7)$          & 34 & 0  & 34 & & +4.0 \\
    $(0.6,0.8)$          & 39 & 0  & 39 & & +4.2 \\
    \midrule
    \multicolumn{6}{@{}l}{\emph{topping up with non-comp blocks}} \\
    $(0.4,0.6)$          & 28 & 24 & 52 & & +3.8 \\
    $(0.5,0.7)$          & 34 & 19 & 53 & & +3.9 \\
    $(0.6,0.8)$          & 39 & 15 & 55 & & +3.7 \\
    \bottomrule
  \end{tabular}
\end{table}

\paragraph{Page-level budget.}
Our policy masks $34\%$ of the page at a band of $(0.5,0.7)$, while a random draw masks around $50\%$. The policy has two ways to increase the total masking budget, and \cref{tab:ladder} shows neither works. One widens the band: the spend on components rises, but under a natural ceiling---components are a subset of the page, and even $(0.6,0.8)$ brings the total to only $39\%$. The other tops up with blocks drawn on the non-component area: no ceiling there, and the total reaches the random budget at $52$--$55\%$. The three bands differ by at most $0.6$ points, and the top-up buys nothing. Within this range, how much of the page is masked matters less than whether every component is.

\begin{figure*}[t]
  \centering
  \includegraphics[width=0.75\textwidth]{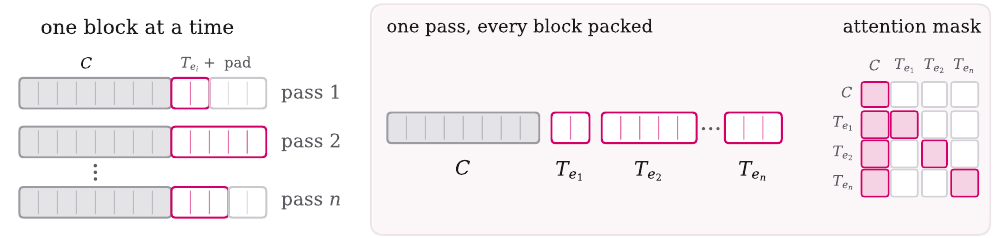}
  \caption{\textbf{Efficient training: all blocks in one pass.} \emph{Left}: one block at a time, as I-JEPA does---$C$ copied and re-encoded on every pass, and blocks padded to the longest in the batch (hatched). The more blocks, the more copies and the more compute. \emph{Middle, right}: we pack all the blocks into one sequence, and an attention mask lets each $T_{e_i}$ see $C$ and itself but never another block. $C$ is then encoded once, the padding goes, and cost follows the patch count rather than the block count.}
  \label{fig:packing}
\end{figure*}

\paragraph{Class-balanced loss.}
Our policy's gain concentrates in the thinly covered classes (\cref{tab:mechanism}), which also contribute the fewest patches to the loss: text appears on
$93\%$ of pages at a median $2.0\%$ of the page, hamburger on $6\%$ at $0.4\%$
(\cref{tab:inventory}).  We therefore ask whether the resulting imbalance can be
corrected through the loss, without increasing mask coverage. We balance the loss so that no class counts for more than
another, and apply it to the random draw as well as to ours. On average,
balancing lifts the \emph{random}$^{+}$ draw further than our mask does, $+2.0$
macro F1 against $+1.3$ (\cref{tab:oursplus}), but the two gains fall on
different classes. On the thinly covered ones ours is ahead, $+3.2$ against
$+1.8$, and higher on all five; on the well covered, balancing is ahead, $+2.1$
against $+0.0$. The two are complementary, and together---our masking with the
balanced loss---they give the largest gain, $+3.7$ macro F1, best in nine of
the twelve classes.

\begin{table}[!t]
  \small
  \centering
  \caption{\textbf{Class-balanced loss.} \emph{+bal}: the loss re-weighted so that every
class's masked patches carry the same total weight. First column:
\emph{random}$^{+}$ (plain loss), absolute test F1; the rest are changes from
it, means over three pretraining seeds, computed before rounding, so they can differ from \cref{tab:mechanism} by
$0.1$. Classes are ordered as in \cref{tab:mechanism}, thinly covered first; bold
marks the best gain in each row.}
  \label{tab:oursplus}
  \setlength{\tabcolsep}{7pt}
  \begin{tabular}{@{}lrrrr@{}}
    \toprule
  & \emph{random}$^{+}$ & \multicolumn{3}{c}{$\Delta$\,F1 over \emph{random}$^{+}$} \\
  \cmidrule(lr){3-5}
  class & F1 & rnd$^{+}$+bal & ours & ours+bal \\
  \midrule
  \multicolumn{2}{@{}l}{\emph{Thinly covered}} & +1.8 & +3.2 & \textbf{+5.2} \\
  \addlinespace[1pt]
  hamburger          & 61.6 & +0.8 & +2.0 & \textbf{+2.5} \\
  login-social-bar   & 57.9 & +2.8 & +6.2 & \textbf{+10.5} \\
  header             & 82.5 & +1.2 & +2.8 & \textbf{+3.9} \\
  menu               & 72.5 & +1.7 & +2.2 & \textbf{+4.0} \\
  dropdown-menu      & 64.5 & +2.6 & +2.7 & \textbf{+5.3} \\
  \midrule
  \multicolumn{2}{@{}l}{\emph{Well covered}} & +2.1 & +0.0 & \textbf{+2.5} \\
  \addlinespace[1pt]
  text-marquee       & 41.6 & \textbf{+7.6} & -1.1 & +6.6 \\
  input-field        & 77.9 & +2.9 & -0.3 & \textbf{+4.8} \\
  button             & 80.0 & +1.1 & +0.9 & \textbf{+1.7} \\
  vector-image       & 55.1 & +1.4 & +1.7 & \textbf{+3.2} \\
  repeater           & 22.6 & \textbf{+0.5} & -0.8 & +0.3 \\
  image              & 73.8 & \textbf{+0.7} & -0.4 & +0.5 \\
  text               & 85.6 & +0.4 & +0.3 & \textbf{+0.7} \\
  \midrule
  macro F1         & 64.6 & +2.0 & +1.3 & \textbf{+3.7} \\
  \bottomrule
  \end{tabular}
\end{table}

\paragraph{Efficient training.}
Reconstructing one block at a time, as I-JEPA does, copies and re-encodes the
whole context on every pass. That is cheap at four blocks a page. Our policy
gives one block per component---fifteen on an average page, forty on a busy one,
and no two pages alike---so the same scheme would copy the context forty times
and pad each batch out to its longest block. We reconstruct all of them in a
single pass instead, following CrossMAE's cross-attention decoding~\cite{fu2024crossmae}:
one sequence per page, with an attention mask that keeps the blocks apart
(\cref{fig:packing}). This costs one departure from I-JEPA's predictor, and it is the departure that
buys the saving. There the predictor runs full self-attention over the context
and the queries together, so the context attends to the queries as well. Here it
does not: the context attends only within itself, and each $T_{e_i}$ attends to
the whole context and to the queries of its own block---cross-attention onto the
context, self-attention inside the block. The context representation is therefore
the same whichever block is being predicted, so it is computed once for the page
instead of once per block. What a block is predicted \emph{from} is unchanged, and the predictor is
identical in every run, so the change cannot favour one mask over another. A page then
costs what its sequence is long---the context once, never copied, plus one query
per masked patch. A random draw
would barely notice the packing; masking every component could not be trained
without it. One run takes about $4.4$ hours on
four L40S GPUs, no more than a random draw costs.

\section{Related Work}
\label{sec:related}

\paragraph{Learning UI representations.}
Many works learn user-interface representations from screenshots and the structured document behind them---the DOM~\cite{whatwg2026dom} on the web, the view hierarchy~\cite{android2026views} on mobile. Rico~\cite{deka2017rico} first paired tens of thousands of mobile screens with their view hierarchies; WebUI~\cite{wu2023webui} later paired webpage screenshots with their DOM at web scale---one vocabulary per site, since each names its own nodes. Screen2Vec~\cite{li2021screen2vec} embeds a Rico screen from view-hierarchy content and interaction traces, never reading the pixels. ActionBert~\cite{he2021actionbert} adds the screenshot, pretraining on large-scale interaction traces, and UIBert~\cite{bai2021uibert} pretrains on both together, masking components of each---image crops, OCR text, hierarchy entries---and predicting them back. Others dropped the structured document, which is noisy in the data and often missing at inference~\cite{li2023spotlight}, and learned from human descriptions of the screen instead: Screen2Words~\cite{wang2021screen2words} captions the whole screen, Spotlight~\cite{li2023spotlight} the widget in focus, and Foresight~\cite{burns2024foresight} the screen an action leads to. Lexi~\cite{banerjee2022lexi} and Screen Correspondence~\cite{wu2023correspondence} replace the human annotation with a hand-built element detector, pretraining by hiding one detected element and predicting its class or text from the features of the rest. Whatever these works hide or predict---structured document features, detected elements or captions---none of them masks the screenshot itself. Two works do. UI-JEPA~\cite{fu2024uijepa} uses no source at all---no document, no descriptions, no detector: it hides random blocks of a UI video and predicts the hidden features from what remains. Pix2Struct~\cite{lee2023pix2struct} hides parts of a webpage screenshot and, from the pixels that remain, predicts the page's simplified HTML---the hidden region is recovered as markup, never as visual content.

\paragraph{Masking strategies.}
Hide the image, predict the image: masked image modeling makes the hidden part itself the target. Two families dominate: MAE~\cite{he2022mae} predicts it in pixel space, I-JEPA~\cite{assran2023ijepa} in latent space. The choice left is where in the image the mask falls, and the choices span a spectrum from blind to authored. The common choice is blind: scattered patches~\cite{he2022mae}, a few blocks~\cite{assran2023ijepa}. Others aim the mask at the image's parts, which must first be found: SemMAE~\cite{li2022semmae} estimates them with a part-segmentation model, AttMask~\cite{kakogeorgiou2022attmask} reads them from a teacher's attention. DSeq-JEPA~\cite{he2026dseqjepa} similarly derives regions from attention, then ranks and predicts them sequentially. ADIOS~\cite{shi2022adios} learns the mask, adversarially against the encoder; HPM~\cite{wang2023hpm} mines the patches hardest to predict; evolved masking~\cite{feng2025evolved} re-draws it as the model's understanding improves. Nearest the authored end, the parts come given, and are masked whole: ObjMAE~\cite{wu2022objmae} masks human-annotated COCO instances to cut pretraining compute; Causal-JEPA~\cite{nam2026causaljepa} masks objects that slot attention finds in video; document pretraining masks the text lines OCR locates~\cite{yu2023structextv2,bai2023wukong}. ODIS~\cite{hizli2025odis} uses provided or detected object masks to align teacher--student crops and constrain object-level attention during self-distillation, while retaining random-block masked patch prediction. In each of these works, the parts are estimated, learned, or annotated. A user interface needs none of this: its structured document already records them.
\paragraph{Mask pretraining on UI.}
Placing a mask from a structured document is not new, and neither is predicting
a hidden region back; we combine the two on a rendered page.
UIBert~\cite{bai2021uibert} masks components its hierarchy locates, but works on
cropped components rather than a page. StrucTexTv2~\cite{yu2023structextv2}
comes closest, masking text regions of a rendered document and regressing their
pixels. Pix2Struct~\cite{lee2023pix2struct} masks the screenshot, but predicts
markup. The mechanism is shared with ObjMAE~\cite{wu2022objmae}, which masks
annotated objects in natural images, and StrucTexTv2, which masks the text a
detector finds in documents; we apply it to UI screenshots, where the page's own
structure (DOM) supplies every component class we keep, and the image encoder
learns by predicting the hidden visual content.

\section{Conclusion}
\label{sec:conclusion}

In this paper we let the page's own structure decide where the mask falls
during pretraining. A rendered page records the components it is built from and
the box each one occupies, so the mask can take part of every component without
a detector to train or a box to draw. Continued pretraining under this policy
raises component macro F1 from $61.5$ to $66.0$ and improves all twelve classes
we evaluate, which neither random draw does.

The gain is not spread evenly, and it was not meant to be. A random draw covers
five of the twelve classes far less than the other seven
(\cref{tab:mechanism}). Ours hides more than half of every class's area, and
the classes that gain are the ones that were short: those five account for the whole of the $1.3$ points we add over a
size-matched random draw. Coverage is measured by running the mask over the
corpus, not by training on it, so which classes a policy starves is known in
advance. We expect, but have not tested, that other domains that record their
parts could supply the same signal; on this corpus the gain falls on the parts a
random mask leaves short.

\section{Limitations}
\label{sec:limitations}

We mask the twelve component classes listed in
\cref{tab:inventory}, which cover $51\%$ of the component instances a page
carries; everything outside that whitelist is treated as non-component area.
This set might not be optimal, and a finer or broader vocabulary may place the
mask better. The same twelve classes also define the evaluation, so the probe
measures the encoder on the components the mask was built from, pooled over
their ground-truth boxes. All pages come from one platform, and all runs use one
backbone and one objective. With no downstream task or external dataset, whether
the gain transfers beyond this measurement is untested, and we do not claim that
it does. Both baselines are random draws: they show that our mask
outperforms a random one on this probe. Our analysis shows that the gain
correlates with coverage (\cref{tab:mechanism}), and that also masking
non-component area adds nothing (\cref{tab:ladder}); other factors---blocks
shaped to each component, never hiding a component whole, and masking
non-component area only---remain to be measured. The mask could also be placed
by estimating the page's parts from the pixels, e.g., with a segmentation model
or a teacher's attention; whether an estimated decomposition can match an
authored one is left open.

{
    \small
    \bibliographystyle{ieeenat_fullname}
    \bibliography{main}
}

\clearpage
\appendix
\crefalias{section}{appendix}
\crefalias{subsection}{appendix}

\makeatletter
\newcommand{\tabinput}[1]{\input{#1} }
\makeatother

\section{Implementation details}
\label{sec:appendix:impl}

\paragraph{Continued pretraining.}
Every arm starts from the public \siglip{} checkpoint and is trained with the same recipe; only the
mask differs. The image tower is NaFlex, so the patch grid is per page, aspect preserving, and
capped at $1{,}024$ patches of $16$ pixels. The I-JEPA predictor is a $6$-layer transformer of
width $384$ with $6$ heads, initialised from scratch. Targets come from the frozen public
checkpoint rather than an EMA teacher, so the prediction target does not drift and the arms remain
comparable to the encoder they start from.

We train for $17{,}000$ steps at an effective batch of $64$ pages ($8$ per device, gradient
accumulation $2$, four GPUs), which is $9.2$ passes over the pretraining pages. The optimiser is
AdamW at a peak learning rate of $5\times10^{-5}$ with a cosine schedule, $1{,}000$ warmup steps,
weight decay $0.05$ and gradients clipped at $1.0$, in \texttt{bf16}. The encoder's learning rate is
held at zero for the first $500$ steps while the freshly initialised predictor warms up: without
this the random predictor drives a large gradient into the encoder in the first few hundred steps
and every arm loses several points of probe accuracy that it does not fully recover. One run takes
about $4.4$ hours on four L40S GPUs.

Pretraining uses the $125{,}277$ training pages, filtered to those whose whitelisted components
cover at least $10\%$ of the patch grid, which leaves $118{,}583$. The filter is applied identically
to every arm, including the random ones, so it cannot favour the component-conditioned policy; its
purpose is to drop pages that are almost entirely background, where no masking policy has anything
to choose between.

\paragraph{The three masking policies.}
All three draw on the same per-page patch grid and all three are re-drawn every epoch.

\emph{Random} samples $4$ blocks, each covering $15$--$20\%$ of the grid with aspect ratio in
$[0.75, 1.5]$. \emph{Random}$^{+}$ samples $16$ blocks of $2$--$8\%$ each, a comparable total budget
spread over more and smaller pieces. Neither reads the page's components. Context is the whole grid
minus the masked union in both cases.

\emph{Ours} first computes the same budget a \emph{random}$^{+}$ draw would spend on this page --
the union of $16$ blocks of $2$--$8\%$ -- and then spends it on the components instead. Each
component instance draws a fraction $\sim\mathcal{U}(0.50, 0.70)$ of itself and hides that much as
one compact slab. Instances are served smallest first, and each quota is a deficit against the
running mask, so a component nested inside another never exceeds its own draw. Two rules protect
the remainder: a component's own slab is capped short of its last visible patch, and a later slab
may not touch the surviving remainder of any component already served, so each component keeps
exactly what its own draw left it. The cap has one boundary effect: an instance the patch grid
resolves to a single patch keeps it and is never masked---about $0.8\%$ of instances. Nothing is
added afterwards to reach the random budget -- the leftover is simply not spent, which is why our
policy masks $34\%$ of the page where a random draw masks around $50\%$.

\paragraph{Single-pass packing.}
The predictor sees one sequence per page, $[\,C, T_{e_1}, \dots, T_{e_n}\,]$: the context patches
followed by one query token per masked patch of each component, each query carrying the sinusoidal
position of the patch it predicts. A per-query-row attention mask does the work that separate
forward passes used to: $C$ attends only within itself, and each $T_{e_i}$ attends to $C$ and to its
own block and to nothing else. Because $C$ never attends to a query, its representation is
block-independent and is therefore computed once rather than once per block, and the per-block
padding disappears -- see \cref{fig:packing}\inmain.

\paragraph{The linear probe.}
A component's feature is the plain mean of the encoder's output over the patches its box touches,
with the box snapped to the grid by the same routine the masking policy uses, so the probe and the
mask agree on what a component's patches are. The probe is
\texttt{StandardScaler} $\rightarrow$ \texttt{LogisticRegression} (lbfgs, \texttt{max\_iter}
$2{,}000$) fitted as one pipeline, so the scaler never sees held-out rows. Standardisation is not
cosmetic here: on raw mean-pooled features a fixed $C$ is not the same amount of regularisation for
two checkpoints whose feature norms differ, and the norms do drift between arms.

The probe trains on the seed split, which is held out of pretraining. Each class contributes
$\min(\text{count}, 1000)$ unique instances: a class with more than $1{,}000$ is page-stratified at
no more than $5$ per page, and a smaller class is taken whole. Nothing is resampled; balance comes
from \texttt{class\_weight='balanced'} rather than from duplicated rows. $C$ is chosen per encoder
from $\{10^{-4}, 10^{-3}, 10^{-2}, 10^{-1}, 1, 10\}$ by $3$-fold cross-validation within these
training instances. The folds are drawn once, grouped by
page (all instances of a page share a fold, so near-duplicate components never sit on both sides)
and stratified by class, and every $C$ is scored on the same folds: fitted on two folds and scored
on the third by balanced accuracy, the mean of per-class recall. The $C$ with the highest mean over
the three folds is kept, ties going to the smaller $C$, and the probe is refitted with it on all
training instances. The grid is the same for every encoder, and nothing is tuned per arm.
Checkpoint curves are read on validation; test is scored once, at the final checkpoint, per arm.
The reported spread is over \emph{pretraining} seeds -- three per arm -- with the probe's own
subsample seed fixed, so the subsampled instances and the cross-validation folds are identical
across encoders and the comparison is paired.

\paragraph{The class-balanced loss (ablation).}
The balanced loss of \cref{tab:oursplus} re-weights the masked patches by class. Let $L_c$ be the
mean loss over the masked patches of class $c$ in a batch and $s_c$ that class's share of those
patches; the plain loss is the share-weighted mixture $\mathcal{L}=\sum_c s_c L_c$, and balancing
changes only the mixture weights:
\begin{equation}
  \mathcal{L}_\alpha \;=\; \sum_{c} w_c\, L_c,
  \qquad w_c \;\propto\; s_c^{\,1-\alpha},
  \qquad \sum_{c} w_c \;=\; 1.
  \label{eq:alpha-mix}
\end{equation}
At $\alpha{=}0$ this is the plain loss; at $\alpha{=}1$ every class present in the batch carries
the same total weight, however many patches it contributes; \cref{tab:oursplus}'s \emph{+bal}
columns train at $\alpha{=}1$. Per patch, each masked patch carries $a_i \propto
s_{c(i)}^{-\alpha}$, normalised so the mean weight over labelled patches is one, which keeps the
loss scale comparable across settings. When components overlap, the smallest owns the patch;
patches on no whitelisted component are background, keep unit weight, and stay out of the share
calculation, so $\alpha{=}0$ reduces to the plain loss with background included. Weights are
computed per batch, with no corpus-level statistics. Balancing needs each masked patch's class;
placing the mask needs only the boxes.

\paragraph{Measuring coverage.}
Coverage is a property of the mask, not of a trained model: for a class, the share of its patches
that the mask hides. We measure it by running each policy over the training pages and counting,
with no encoder in the loop, which is why the coverage a policy will give each class is known
before any pretraining is done.

\section{The dataset}
\label{sec:appendix:data}

\subsection{What we release}
\label{sec:release}

We release the data used by our frozen-encoder linear-probe evaluation as a
Hugging Face dataset with two splits: a \texttt{seed} split of $4{,}000$ pages
and a held-out \texttt{test} split of $5{,}000$ pages. The seed split is used to
fit the probe and select its regularization by grouped cross-validation; the
test split is reserved for final evaluation. Together, the release contains
$9{,}000$ pages. \Cref{tab:appendix:splitcounts} reports the per-split instance
counts for the twelve component classes.

Each row represents one rendered homepage and contains exactly three fields:
\texttt{image}, the screenshot itself; \texttt{classes}, an ordered list of
component-class strings; and \texttt{rectangles}, the corresponding ordered
list of bounding boxes. Thus, \texttt{classes[i]} labels
\texttt{rectangles[i]}. Boxes use absolute page-pixel coordinates in
$(x_{\min}, y_{\min}, x_{\max}, y_{\max})$ order, with the origin at the
top-left of the screenshot. An abbreviated record has the following form:

{\footnotesize
\begin{verbatim}
{
  "image": <PIL.Image.Image>,
  "classes": ["header", "menu", "button", ...],
  "rectangles": [
    [x_min, y_min, x_max, y_max],
    [x_min, y_min, x_max, y_max],
    ...
  ]
}
\end{verbatim}
}

The screenshot bytes are embedded directly in the dataset. The released rows
contain no S3 references, website URLs or domains, crawl-source fields, DOM
text, or private internal identifiers. The class strings are included in each
record, while the dataset card specifies the complete twelve-class vocabulary
and its canonical spelling.

The dataset can be loaded with the Hugging Face \texttt{datasets} library:

{\footnotesize
\begin{verbatim}
from datasets import load_dataset

dataset = load_dataset("Wix/ui-components")
seed = dataset["seed"]
test = dataset["test"]

example = seed[0]
image = example["image"]
classes = example["classes"]
rectangles = example["rectangles"]
\end{verbatim}
}

\paragraph{Ethics statement.}
The released seed and test splits contain screenshots captured from the
publicly accessible rendered homepages of sites created on our platform; the
collection accessed no authenticated or private user data, and any personal
information inadvertently included in a screenshot was content already made
publicly available by the site publisher at the time of collection. The release
carries a takedown mechanism: a site owner can request the removal of their page
through the dataset's Hugging Face page, and removed pages leave the release in
versioned updates. Each version's page list is recorded on the dataset card, so
any reported number is pinned to a version and can be recomputed on any later
one; at $5{,}000$ test pages, each removal changes the evaluation set by
$0.02\%$. The study analyzes publicly available web content and involves
no interaction with human subjects, and therefore did not require institutional
review.

\begin{table}[t]
  \centering
  \small
  \caption{Component instances per split, on the same basis as \cref{tab:inventory} and
  \cref{tab:appendix:vocab-kept}: the eligibility-gated harvest. The columns sum to the
  $2{,}131$K whitelisted instances of \cref{tab:inventory}. The seed split trains the probe,
  validation carries the checkpoint curves, and test is read once. Classes ordered as in
  \cref{tab:mechanism}, thinly covered first.}
  \label{tab:appendix:splitcounts}
  \setlength{\tabcolsep}{5pt}
  \begin{tabular}{@{}lrrrr@{}}
    \toprule
    class & train & seed & val & test \\
    \midrule
    hamburger        & 7{,}808   & 266    & 115    & 319 \\
    login-social-bar & 14{,}650  & 474    & 229    & 523 \\
    header           & 107{,}681 & 3{,}461 & 1{,}573 & 3{,}908 \\
    menu             & 71{,}058  & 2{,}200 & 1{,}038 & 2{,}540 \\
    dropdown-menu    & 23{,}914  & 778    & 357    & 861 \\
    \midrule
    text-marquee     & 5{,}987   & 212    & 104    & 229 \\
    input-field      & 14{,}315  & 513    & 199    & 513 \\
    button           & 236{,}265 & 7{,}572 & 3{,}572 & 8{,}657 \\
    vector-image     & 137{,}943 & 4{,}547 & 2{,}025 & 4{,}907 \\
    repeater         & 34{,}738  & 1{,}072 & 512    & 1{,}274 \\
    image            & 396{,}702 & 12{,}544 & 5{,}967 & 14{,}580 \\
    text             & 915{,}578 & 29{,}213 & 13{,}302 & 34{,}206 \\
    \midrule
    \emph{pages}     & 125{,}277 & 4{,}000 & 2{,}000 & 5{,}000 \\
    \bottomrule
  \end{tabular}
\end{table}

\subsection{The full component vocabulary}
\label{sec:appendix:taxonomy}
\cref{tab:inventory}\inmain{} lists the seventeen platform tokens we keep and folds
everything else into a single row. \cref{tab:appendix:vocab-kept} gives the same seventeen with
their exact instance counts, and \cref{tab:appendix:vocab-ignored} opens up the row we folded: the
tokens the crawl saw and the taxonomy ignores, ordered by frequency. Counts are on the same basis as
\cref{tab:inventory} -- after the eligibility gate, which drops occluded, off-canvas and very small
instances -- so the two tables can be read together.

Most of what we ignore is structural. \texttt{box} and \texttt{section} alone are the two largest
tokens in the corpus and they arrange other components rather than carrying content, which is the
reason a component is defined by its platform token and not by being a leaf of the DOM.

\begin{table}[t]
  \centering
  \scriptsize
  \caption{The seventeen kept tokens and the twelve classes they map to, with the share of pages
  carrying each token and its eligible instance count. Same basis as \cref{tab:inventory}.}
  \label{tab:appendix:vocab-kept}
  \setlength{\tabcolsep}{4pt}
  \begin{tabular}{@{}llrr@{}}
    \toprule
    token & class & pages & instances \\
    \midrule
        \texttt{rich-text} & text & 93\% & 989,569 \\
    \texttt{image} & image & 84\% & 405,800 \\
    \texttt{button} & button & 69\% & 256,087 \\
    \texttt{vector-image} & vector-image & 42\% & 149,439 \\
    \texttt{header} & header & 85\% & 116,627 \\
    \texttt{horizontal-menu} & menu & 54\% & 75,716 \\
    \texttt{repeater} & repeater & 22\% & 37,597 \\
    \texttt{dropdown-menu} & dropdown-menu & 19\% & 25,910 \\
    \texttt{video-box} & image & 11\% & 18,821 \\
    \texttt{login-social-bar} & login-social-bar & 12\% & 15,878 \\
    \texttt{text-input} & input-field & 5\% & 11,501 \\
    \texttt{hamburger-open-button} & hamburger & 6\% & 8,508 \\
    \texttt{text-marquee} & text-marquee & 4\% & 6,532 \\
    \texttt{video-player} & image & 3\% & 5,195 \\
    \texttt{search-bar} & input-field & 3\% & 4,041 \\
    \texttt{collapsible-text} & text & 1\% & 2,762 \\
    \texttt{vertical-menu} & menu & 1\% & 1,126 \\
 
    \bottomrule
  \end{tabular}
\end{table}

\begin{table}[t]
  \centering
  \scriptsize
  \caption{The tokens the taxonomy ignores, most frequent first, with the tail collapsed into the
  final row. These are treated as non-component area of the screenshot: a random draw may land on
  them, and our policy never places a slab on them.}
  \label{tab:appendix:vocab-ignored}
  \setlength{\tabcolsep}{4pt}
  \begin{tabular}{@{}llrr@{}}
    \toprule
    source & token & pages & instances \\
    \midrule
        \texttt{\clspfx} & \texttt{box} & 72\% & 1,369,175 \\
    \texttt{\clspfx} & \texttt{section} & 95\% & 366,707 \\
    \texttt{\clspfx} & \texttt{column-strip} & 20\% & 55,687 \\
    \texttt{bg} & \texttt{bg-image} & 27\% & 45,669 \\
    \texttt{dom} & \texttt{social-icon} & 10\% & 37,157 \\
    \texttt{bg} & \texttt{bg-video} & 19\% & 31,099 \\
    \texttt{tpa} & \texttt{gallery-wrapper-comp-img} & 6\% & 28,410 \\
    \texttt{tpa} & \texttt{gallery-wrapper-comp} & 10\% & 16,167 \\
    \texttt{tpa} & \texttt{form} & 4\% & 10,063 \\
    \texttt{\clspfx} & \texttt{language-menu} & 5\% & 6,717 \\
    \texttt{tpa} & \texttt{pro-gallery-comp-img} & 1\% & 5,627 \\
    \texttt{\clspfx} & \texttt{slideshow} & 2\% & 3,414 \\
    \texttt{\clspfx} & \texttt{form} & 2\% & 3,093 \\
    \texttt{\clspfx} & \texttt{search-button} & 2\% & 2,851 \\
    \texttt{\clspfx} & \texttt{footer} & 2\% & 2,738 \\
    \texttt{\clspfx} & \texttt{gallery} & 2\% & 2,707 \\
    \texttt{\clspfx} & \texttt{multi-state-box} & 1\% & 2,531 \\
    \texttt{\clspfx} & \texttt{image-button} & 1\% & 2,450 \\
    \texttt{tpa} & \texttt{pro-gallery-comp} & 1\% & 2,386 \\
    \texttt{dom} & \texttt{video} & 1\% & 1,509 \\
    \texttt{\clspfx} & \texttt{text-box} & 1\% & 1,279 \\
    \texttt{\clspfx} & \texttt{google-map} & 1\% & 1,217 \\
    \texttt{\clspfx} & \texttt{horizontal-line} & 0\% & 1,159 \\
    \texttt{\clspfx} & \texttt{accordion} & 1\% & 1,091 \\
    \texttt{\clspfx} & \texttt{audio-player} & 0\% & 1,002 \\
    \texttt{tpa} & \texttt{events-widget} & 1\% & 821 \\
    \texttt{\clspfx} & \texttt{tabs} & 0\% & 696 \\
    \texttt{\clspfx} & \texttt{dropdown} & 0\% & 582 \\
    \texttt{\clspfx} & \texttt{checkbox} & 0\% & 412 \\
    \texttt{\clspfx} & \texttt{anchor-menu} & 0\% & 389 \\
    \texttt{\clspfx} & \texttt{text-effects} & 0\% & 305 \\
    \texttt{\clspfx} & \texttt{lottie-embed} & 0\% & 300 \\
    \texttt{\clspfx} & \texttt{vertical-line} & 0\% & 237 \\
    \texttt{tpa} & \texttt{path-play} & 0\% & 186 \\
    \texttt{tpa} & \texttt{service-info-aria-section-title} & 0\% & 180 \\
    \texttt{\clspfx} & \texttt{text-on-path} & 0\% & 164 \\
    \texttt{tpa} & \texttt{new-blog-popover-root} & 0\% & 147 \\
    \texttt{\clspfx} & \texttt{ratings-display} & 0\% & 127 \\
    \texttt{\clspfx} & \texttt{date-picker} & 0\% & 125 \\
    \texttt{tpa} & \texttt{label-for-id} & 0\% & 120 \\
    \multicolumn{2}{@{}l}{\emph{128 further tokens}} & --- & 1,552 \\
 
    \bottomrule
  \end{tabular}
\end{table}

\begin{figure*}[t]
  \centering
  \includegraphics[width=0.74\textwidth]{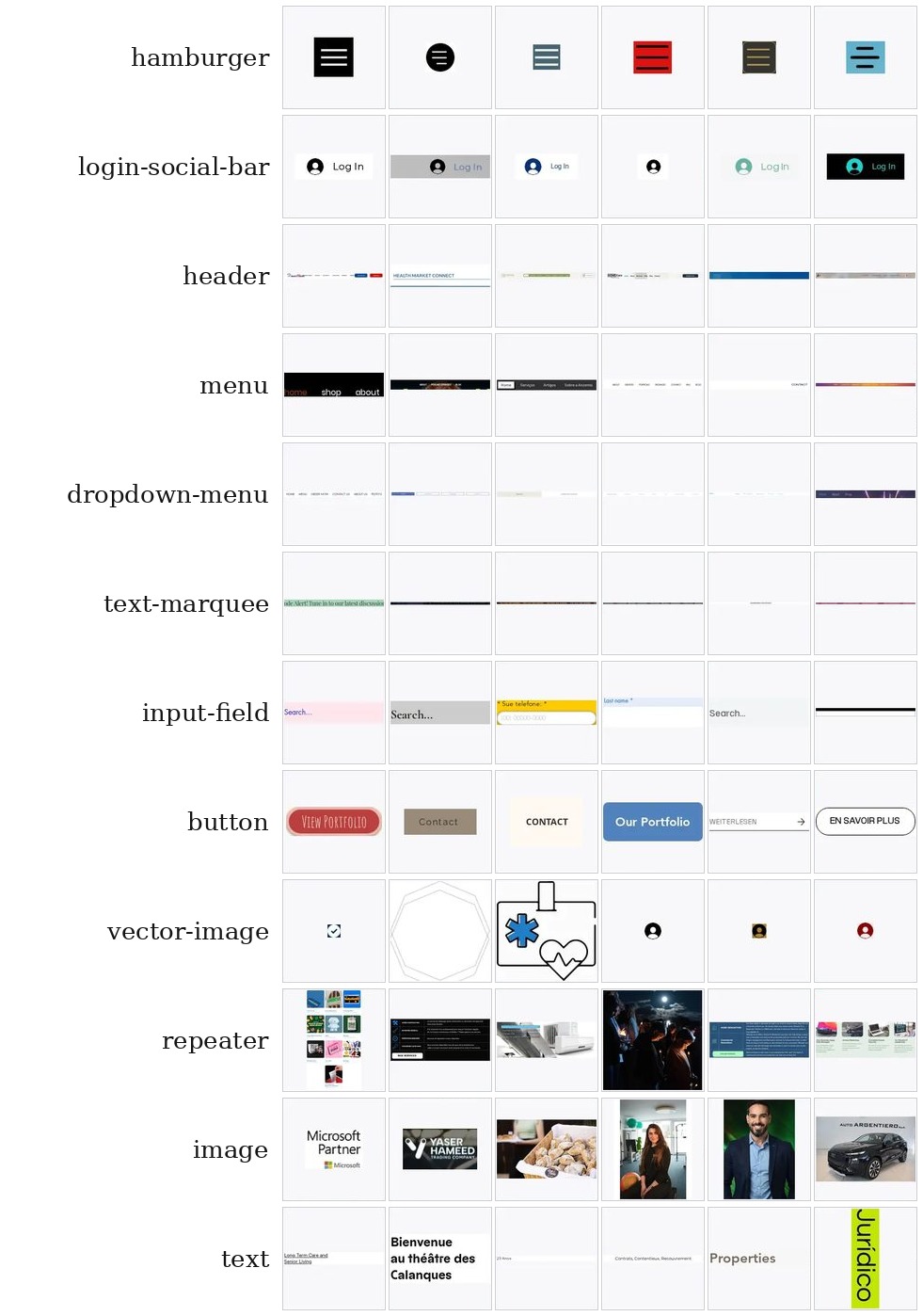}
  \caption{Six instances of each class, cropped from the pages they appear on and shown at a common
  cell size, so wide components such as \emph{header} and \emph{menu} appear as slivers of their
  cell while small ones fill it. Worth looking at before reading any of the numbers: a
  \emph{login-social-bar} is a login bar on every page it appears on, and it is black, white, pink
  or photographic, with an icon or without, in half a dozen languages. A class is defined by what
  the platform says it is, not by how it looks, which is what makes recovering it a representation
  problem rather than a template-matching one.}
  \label{fig:appendix:gallery}
\end{figure*}

\subsection{Screenshot dimensions}
\label{sec:appendix:shots}
Pages are rendered at $1280$ pixels wide and cut at the section boundary nearest a $2{,}160$ pixel
target, so height varies and width does not: $96\%$ of pages are exactly $1280$ wide. Heights run
from $1{,}000$ to $12{,}000$ pixels with a median of $1{,}877$; the quartiles are $1{,}586$ and
$2{,}105$, and the $5$th and $95$th percentiles $1{,}146$ and $2{,}562$. A third of pages land
within $10\%$ of the target and $20.5\%$ overshoot it, which is the cost of cutting at a boundary
rather than at a fixed line: a section that runs long carries the cut past the target rather than
splitting a component. In aspect ratio the median page is $1.46$ times as tall as it is wide, with
the middle half between $1.23$ and $1.64$.

\section{Full per-class results}
\label{sec:appendix:perclass}

\subsection{Per-class F1}
\cref{tab:headline}\inmain{} reports three-seed means on the test split. The two tables
here give what those means were computed from: every class, every arm, on validation
(\cref{tab:appendix:perclass-val}) and on test (\cref{tab:appendix:perclass-test}), with the spread
over the three pretraining seeds. Validation is the split the protocol reads freely; test is read
once. The two agree on both group-level facts the paper rests on -- ours gains most on the thinly
covered classes, and ours and \emph{random}$^{+}$ are level on the well covered ones.

\begin{table}[t]
  \centering
  \small
  \caption{Per-class F1 on the \textbf{validation} split. Frozen is a single encoder; each
  continued-pretrained arm is the mean over three pretraining seeds, $\pm$ their standard
  deviation. Classes ordered as in \cref{tab:mechanism}, thinly covered first.}
  \label{tab:appendix:perclass-val}
  \setlength{\tabcolsep}{4pt}
  \begin{tabular}{@{}lrrrr@{}}
    \toprule
    class & frozen & random & random$^{+}$ & ours \\
    \midrule
        hamburger & 63.4 & 57.5\,\tiny{$\pm$\,1.6} & 59.0\,\tiny{$\pm$\,1.4} & 63.6\,\tiny{$\pm$\,1.2} \\
    login-social-bar & 63.9 & 57.3\,\tiny{$\pm$\,1.6} & 57.9\,\tiny{$\pm$\,0.5} & 66.5\,\tiny{$\pm$\,1.4} \\
    header & 78.4 & 81.7\,\tiny{$\pm$\,0.1} & 82.5\,\tiny{$\pm$\,0.3} & 85.1\,\tiny{$\pm$\,0.5} \\
    menu & 68.5 & 71.8\,\tiny{$\pm$\,2.0} & 73.2\,\tiny{$\pm$\,0.9} & 75.2\,\tiny{$\pm$\,0.8} \\
    dropdown-menu & 52.4 & 63.8\,\tiny{$\pm$\,1.2} & 62.7\,\tiny{$\pm$\,0.5} & 66.0\,\tiny{$\pm$\,0.6} \\
    text-marquee & 41.6 & 41.8\,\tiny{$\pm$\,2.2} & 43.4\,\tiny{$\pm$\,2.0} & 41.8\,\tiny{$\pm$\,1.7} \\
    input-field & 76.0 & 75.9\,\tiny{$\pm$\,1.6} & 78.2\,\tiny{$\pm$\,1.0} & 80.6\,\tiny{$\pm$\,1.8} \\
    button & 79.6 & 80.1\,\tiny{$\pm$\,0.1} & 81.4\,\tiny{$\pm$\,0.6} & 82.1\,\tiny{$\pm$\,1.0} \\
    vector-image & 56.2 & 56.2\,\tiny{$\pm$\,0.1} & 57.4\,\tiny{$\pm$\,0.3} & 58.7\,\tiny{$\pm$\,0.6} \\
    repeater & 18.6 & 21.7\,\tiny{$\pm$\,0.6} & 21.9\,\tiny{$\pm$\,0.1} & 21.3\,\tiny{$\pm$\,0.5} \\
    image & 69.6 & 71.8\,\tiny{$\pm$\,0.4} & 72.5\,\tiny{$\pm$\,0.2} & 72.0\,\tiny{$\pm$\,0.6} \\
    text & 82.5 & 84.6\,\tiny{$\pm$\,0.4} & 85.4\,\tiny{$\pm$\,0.2} & 85.9\,\tiny{$\pm$\,0.3} \\
    \midrule
    macro F1 & 62.6 & 63.7\,\tiny{$\pm$\,0.6} & 64.6\,\tiny{$\pm$\,0.3} & 66.6\,\tiny{$\pm$\,0.5} \\
 
    \bottomrule
  \end{tabular}
\end{table}

\begin{table}[t]
  \centering
  \small
  \caption{Per-class F1 on the sealed \textbf{test} split, read once per arm at the final
  checkpoint. \cref{tab:headline}\inmain{} reports these as changes from frozen.}
  \label{tab:appendix:perclass-test}
  \setlength{\tabcolsep}{4pt}
  \begin{tabular}{@{}lrrrr@{}}
    \toprule
    class & frozen & random & random$^{+}$ & ours \\
    \midrule
        hamburger & 62.5 & 60.0\,\tiny{$\pm$\,1.4} & 61.6\,\tiny{$\pm$\,2.3} & 63.7\,\tiny{$\pm$\,0.8} \\
    login-social-bar & 62.1 & 55.4\,\tiny{$\pm$\,0.7} & 57.9\,\tiny{$\pm$\,0.8} & 64.1\,\tiny{$\pm$\,1.0} \\
    header & 77.8 & 82.0\,\tiny{$\pm$\,0.5} & 82.5\,\tiny{$\pm$\,0.7} & 85.3\,\tiny{$\pm$\,0.3} \\
    menu & 65.6 & 70.5\,\tiny{$\pm$\,1.2} & 72.6\,\tiny{$\pm$\,0.9} & 74.7\,\tiny{$\pm$\,0.2} \\
    dropdown-menu & 53.2 & 64.1\,\tiny{$\pm$\,1.2} & 64.5\,\tiny{$\pm$\,0.6} & 67.2\,\tiny{$\pm$\,0.6} \\
    text-marquee & 35.4 & 39.3\,\tiny{$\pm$\,2.4} & 41.6\,\tiny{$\pm$\,1.7} & 40.5\,\tiny{$\pm$\,1.8} \\
    input-field & 75.9 & 73.6\,\tiny{$\pm$\,2.0} & 77.9\,\tiny{$\pm$\,0.6} & 77.6\,\tiny{$\pm$\,0.7} \\
    button & 78.8 & 78.8\,\tiny{$\pm$\,0.1} & 80.0\,\tiny{$\pm$\,0.2} & 80.9\,\tiny{$\pm$\,0.4} \\
    vector-image & 53.4 & 54.4\,\tiny{$\pm$\,0.4} & 55.1\,\tiny{$\pm$\,0.5} & 56.8\,\tiny{$\pm$\,0.7} \\
    repeater & 19.3 & 22.1\,\tiny{$\pm$\,0.2} & 22.6\,\tiny{$\pm$\,0.4} & 21.8\,\tiny{$\pm$\,0.4} \\
    image & 71.1 & 72.9\,\tiny{$\pm$\,0.6} & 73.8\,\tiny{$\pm$\,0.4} & 73.4\,\tiny{$\pm$\,0.8} \\
    text & 82.7 & 84.8\,\tiny{$\pm$\,0.3} & 85.6\,\tiny{$\pm$\,0.3} & 85.9\,\tiny{$\pm$\,0.2} \\
    \midrule
    macro F1 & 61.5 & 63.1\,\tiny{$\pm$\,0.7} & 64.6\,\tiny{$\pm$\,0.5} & 66.0\,\tiny{$\pm$\,0.2} \\
 
    \bottomrule
  \end{tabular}
\end{table}

\subsection{The $\alpha$ sweep}
\label{sec:appendix:loss}

\paragraph{Purpose of the sweep.}
The parameter $\alpha$ interpolates between the plain objective
($\alpha{=}0$), where class contributions follow their masked-patch
frequencies, and full class balancing ($\alpha{=}1$), where every class
present in the batch receives equal aggregate weight. We use $\alpha{=}1$
because it is the fully balanced endpoint, not because it was selected from
the sweep. Intermediate values are matched single-seed diagnostics showing
how the representations change between these two objectives; the replicated
three-seed endpoint comparisons are reported in \cref{tab:oursplus}.

\begin{table}[t]
  \centering
  \small
  \caption{\textbf{Validation macro F1 across the class-balancing interpolation.}
At $\alpha{=}0$ the objective reduces to the plain loss; at $\alpha{=}1$ every
class present in the batch receives equal aggregate weight. Intermediate
settings are matched single-seed diagnostics and were not used to choose the
fully balanced endpoint.}
  \label{tab:appendix:alpha-val}
  \begin{tabular}{@{}lrrr@{}}
  \toprule
  $\alpha$ & random$^{+}$ & ours & gap \\
  \midrule
  $0.00$ & 64.20 & 66.58 & +2.38 \\
  $0.25$ & 65.75 & 67.95 & +2.20 \\
  $0.50$ & 66.00 & 68.19 & +2.19 \\
  $0.75$ & 66.76 & 68.80 & +2.04 \\
  $1.00$ & 66.68 & 68.42 & +1.74 \\
  \bottomrule
\end{tabular}
\end{table}

\end{document}